\documentclass{article} 
\usepackage[english]{babel}
\usepackage{amssymb}
\usepackage{amsmath}
\usepackage{txfonts}
\usepackage{mathdots}
\usepackage[classicReIm]{kpfonts}
\usepackage[dvips]{graphicx} %%% use 'pdftex' instead of 'dvips' for PDF output
\def\BibTeX{{\rm B\kern-.05em{\sc i\kern-.025em b}\kern-.08em
		T\kern-.1667em\lower.7ex\hbox{E}\kern-.125emX}}

\begin{document}

\title {Magnetically guided capsule endoscopy}

\author {Thomas Kruezer}

\maketitle

\begin{abstract} The following research undertakes a historical review of this technology with specific highlighting of its advancement in medical diagnostics as well as the therapeutic functionality of wireless capsule endoscopy. Without restriction to the gastrointestinal tract alone, the review will additionally investigate the developments in the technology of micro-robots guided through the magnetic power and are capable of navigating through multiple forms of air and fluid filled lumina as well as cavities within the body. All these capabilities are of use in the utilization of minimally invasive medicine.
\end{abstract}

\section{Introduction}
\label{sec:introduction}

Even with the reduction in incidence and mortality in gastric related complication cases globally, it remains as the second most deadly digestive neoplasm. Worthy to note is that intestinal metaplasia and dysplasia are precursors to cancer [1]. In this case, identifying these lesions among patients preceded by follow-up of the afflicted areas can provide a means of early diagnosis and treatment. Furthermore, it could, therefore, bring about an enhancement of patient survival [22]. Therefore, wireless capsule endoscopy is a formidable resource used in medical screening and diagnostic procedures. 

It involves the swallowing of a small capsule whose movement in the body is determined by natural peristalsis as well as gravity within the human gastrointestinal tract [2]. This capsule is fitted with a camera which is integrated to allow for the visualization of the small intestine region which in the past would not be accessed using traditional flexible endoscopy [23].

Furthermore, as an endoscopy resource, it allows for the localization of bleeding sources in the midst of the gastrointestinal tract and the identification of diseases, for instance, inflammatory bowel disease otherwise known as Crohn's diseases, tumors as well as polyposis among others [3]. The process of screening and associated diagnostic efficacy of wireless capsule endoscopy more so within the stomach region is limited by multiple technical difficulties [5]. These include, inadequate active capsular positioning as well as orientation control.  

In addition, there is the absence of therapeutic functionality within most commercially available capsules due to limitations of the volume of the capsule and the storage of energy [6]. In this case, the likelihood of the usage of body-exogenous magnetic fields in guiding, orienting, powering and operating the capsule as well as its mechanisms has led to growing interest in research of magnetically guided capsule endoscopy.

\section{BACKGROUND}
\label{sec:background}
Multiple methods are accessible for the enhancement and emphasizing of mucosa irregularities and for augmenting the visibility of structures identified below the surface of the mucosa. The most critical techniques entail narrow band imaging, magnification endoscopy, chromoendoscopy, confocal laser endomicroscopy, and optical coherence tomography as well as magnification endoscopy [22]. These methods have normally been evaluated against each other or integrated on the basis of their effect on diagnostic accuracy. However, a real distinction between chromoendoscopy as well as all competing methods is identified in the absence of augmented hardware [21]. Chromoendoscopy needs no change in hardware of the imaging system itself [7].

\section{CAPSULE ENDOSCOPE DESIGN}
The core concept behind the design of the capsule endoscope has not undergone significant changes over the years. The external shell of the capsule entails a biocompatible polycarbonate material with the half-spherical ends being transparent with surrounded image LED sensors. Adjacently behind the sensor lays the ASIC which is responsible for the processing of signals and radio transmitter unit control. The radio signal normally transmitted is based on the Industrial Scientific and Medical Band channel of 434.79 MHz [8].

Power is availed by a singular or multiple button cells which are compactly fitted within the capsule. Furthermore, the size of commercially available variations of the capsule is normally less than the full dimension of 32 mm in length with a diameter of 13 mm to allow for ease in swallowing [9]. The method of using a single camera does not provide a full guarantee in the process of screening since the orientation control of the capsule cannot be achieved [10]. As such, this challenge can be tackled by introducing a camera with a broader viewing angle for new versions [20]. In addition, capsules with many cameras have also been introduced such as the PillCam COLON used in colonoscopy as well as the PillCam ESO utilized in esophagus examination with the inclusion of double end facing cameras [11].

\section{STATE OF RESEARCH AND DEVELOPMENT}

It is noted that form the year 2010, commercial variations of the wireless capsule endoscopy did not have active propulsion as well as orientation control. In addition, they were mainly applied in targeted use for video-based screening as well as mucosal pathology diagnosis for regions such as the esophagus as well as large and small intestines [12].  The full screening of the gastrointestinal region using traditional gastroscopy could not be achieved. As such, the clinical advent of magnetically guided capsule endoscopy which utilizes magnetic fields as a means of orienting and propelling the capsule marked the beginning for a patient-friendly capsule-based option to gastroscopy [13].

\section{NON-MAGNETIC MINIMALLY INVASIVE TECHNIQUES IN GASTROENTEROLOGY}

\subsection {Active locomotion}

From the start of the clinical gains of wireless capsule endoscopy as a formidable diagnosis resource for the gastrointestinal tract, the formulation of a vigorous momentum system has been the foremost objective of research factions working towards the advancement of this technology [14]. The passive capsular motions, impacts of peristalsis as well as gravity are all factors that have restricted the technology in regard to diagnostic efficacy [24]. In addition, the absence of position, orientation and velocity control has lead to unfulfilled pathological screening [15]. In this case, development of active propulsion methods and associated technological advancements in wireless capsule endoscopy call for a complete understanding of the geometry as well as the biomechanics of the gastrointestinal tract [16].

\subsection{Physiological sensing and therapeutics}

In augmentation to video imaging, wireless capsule endoscopy technology for measuring intraluminal physiological environments can be of extreme usage for gastric functionality monitoring. Disorders for instance gastroparesis as well as chronic constipation can interrupt typical gastric motility [17]. Furthermore, in regard to physiological parameters, for instance, intraluminal, pressure, Ph, temperature, the complete and regional gastrointestinal transit times are vital for measuring diagnosis of such gastric disorders [18].

\section{MAGNETICALLY MINIMAL INVASIVE TECHNIQUES IN GASTROENTEROLOGY}

Magnetic fields externally generated from a patient's body can be put to use in exerting translational and rotational forces on the capsule since it is embedded with permanent magnets or materials that are magnetizable [21]. In this case, active locomotion methods that make the use of shape memory alloys, electrical motor drivers as well as piezo drives otherwise known as internal locomotion techniques require augmented internal power supplies to operate [19].

In addition, they also require external tethering using flexible power cords. Therefore, the usage of magnetic fields of capsule locomotion makes wireless capsule endoscopy realistically wireless based on its external locomotion mechanism. Furthermore, positioning, orientation and capsule velocity control in the augmentation of the magnetic fields could also be harnessed to generate power for the capsule and charge its batteries for localized imaging within the body [20].

\section{WHY IS ODOMETRY IMPORTANT IN ENDOSCOPY}  

According to Turan, it is vital that all forms of modern reconstruction of 3D mobile applications use a modular fashion that comprises of key frame selection, 3D based shading, sparse dense pose alignment and pre-processing. These methods are important in odometry endoscopy because they solve the problem of real time precise localization by allowing for active control of the endoscopic robot capsule.

\section{CONCLUDING REMARKS}

Magnetic capsule endoscopy for usage in diagnostic procedures for regions such as the gut and colon are faced with three main concerns. The first being extensive complications in gastric preparation. Secondly, another concern is the absence of capabilities in biopsy and thirdly the comparatively time-consuming examination process in comparison to traditional gastroscopy or colonoscopy.

As such, the general consensus among clinicians seemingly points to magnetically generated capsule endoscopy offering no specific diagnostic benefits over traditional gastroscopy of the upper gastrointestinal tract. In this case, the efficacy of magnetically generated capsule demands for targeted investigation for individual diagnosis of disorders.


\begin{thebibliography}{99}

	
\bibitem{b1} Armaroli P, Villain P, Suonio E, Almonte M, Anttila A, Atkin WS, Dean PB, de Koning HJ, Dillner L, Herrero R. European code against cancer: cancer screening. Cancer Epidemiol. 2015;39:S139CS152. doi: 10.1016/j.canep.2015.10.021. [PubMed] [Cross Ref]

\bibitem{b2} Hassan C, Rossi PG, Camilloni L, Rex D, Jimenez{\textbackslash}Cendales B, Ferroni E, Borgia P, Zullo A, Guasticchi G, Group H. Meta{\textbackslash}analysis: adherence to colorectal cancer screening and the detection rate for advanced neoplasia, according to the type of screening test. Aliment Pharmacol Ther. 2012;36:929C940. doi: 10.1111/apt.12071. [PubMed] [Cross Ref]

\bibitem{b3} Senore C, Inadomi J, Segnan N, Bellisario C, Hassan C (2015) Optimising colorectal cancer screening acceptance: a review. Gut, gutjnl-[Call] [PubMed]

\bibitem{b4} Leung WC, Foo DC, Chan TT, Chiang MF, Lam AH, Chan HH, Cheung CC (2016) Alternatives to colonoscopy for population-wide colorectal cancer screening. Hong Kong medical journal = Xianggang Yi Xue za zhi / Hong Kong Academy of Medicine [PubMed] 

\bibitem{b5} Trevisani L, Zelante A, Sartori S. Colonoscopy, pain, and fears: is it an indissoluble trinomial? World J Gastrointest Endosc. 2014;6:227C233. doi: 10.4253/wjge.v6.i6.227. [PMC free article] [PubMed][Cross Ref]

\bibitem{b6} Iddan G, Meron G, Glukhovsky A, Swain P. Wireless capsule endoscopy. Nature. 2000;405:417. doi: 10.1038/35013140. [PubMed] [Cross Ref]

\bibitem{b7} Koulaouzidis A, Iakovidis DK, Karargyris A, Rondonotti E. Wireless endoscopy in, Will it still be a capsule? World J Gastroenterol. 2015;2020\eqref{17}:5119C5130. doi: 10.3748/wjg.v21.i17.5119.[PMC free article] [PubMed] [Cross Ref]

\bibitem{b8} Ciuti G, Menciassi A, Dario P. Capsule endoscopy: from current achievements to open challenges. IEEE Rev Biomed Eng. 2011;4:59C72. doi: 10.1109/RBME.2011.2171182. [PubMed] [Cross Ref]

\bibitem{b9} Loeve A, Breedveld P, Dankelman J. Scopes too flexible{\dots} And too stiff. IEEE Pulse. 2010;1:26C41. doi: 10.1109/MPUL.2010.939176. [PubMed] [Cross Ref]

\bibitem{b10} Sliker LJ, Ciuti G. Flexible and capsule endoscopy for screening, diagnosis, and treatment. Expert Rev Med Devices. 2014;11:649C666. doi: 10.1586/17434440.2014.941809. [PubMed] [Cross Ref] 

\bibitem{b11} Turan, M., Almalioglu, Y., Araujo, H., Konukoglu, E., \& Sitti, M. (2017). A non-rigid map fusion-based direct SLAM method for endoscopic capsule robots.~\textit{International journal of intelligent robotics and applications},~\textit{1}(4), 399-409.

\bibitem{b12} M Turan, Y Almalioglu, H Araujo, E Konukoglu, M Sitti. Deep endovo: A recurrent convolutional neural network (rcnn) based visual odometry approach for endoscopic capsule robots. (2016). Neuro computing .275, 1861-1870, 2018

\bibitem{b13} M Turan, YY Pilavci, I Ganiyusufoglu, H Araujo, E Konukoglu, M Sitti. (2018). Sparse-then-dense alignment-based 3D map reconstruction method for endoscopic capsule robots. Machine Vision and Applications.~\textit{29}(2), 345-359

\bibitem{b14} M Turan, J Shabbir, H Araujo, E Konukoglu, M Sitti. (2017).A deep learning based fusion of RGB camera information and magnetic localization information for endoscopic capsule robots. International journal of intelligent robotics and applications {1}(4), 442-450

\bibitem{b15} Ciuti, G., Cali\`{o}, R., Camboni, D., Neri, L., Bianchi, F., Arezzo, A., ... \& Magnani, B. (2016). Frontiers of robotic endoscopic capsules: a review.~\textit{Journal of micro-bio robotics},~\textit{11}(1-4), 1-18.

\bibitem{b16} Hosoe, N., Naganuma, M., \& Ogata, H. (2015). Current status of capsule endoscopy through a whole digestive tract.~\textit{Digestive Endoscopy},~\textit{27}(2), 205-215.

\bibitem{b17} Spyrou, E., \& Iakovidis, D. K. (2013). Video-based measurements for wireless capsule endoscope tracking.~\textit{Measurement Science and Technology},~\textit{25}(1), 015002.

\bibitem{b18} Halling, M. L., Nathan, T., Kjeldsen, J., \& Jensen, M. D. (2014). High sensitivity of quick view capsule endoscopy for detection of small bowel C rohn's disease.~\textit{Journal of gastroenterology and hepatology},~\textit{29}(5), 992-996.

\bibitem{b19} Hejazi, R. A., Bashashati, M., Saadi, M., Mulla, Z. D., Sarosiek, I., McCallum, R. W., \& Zuckerman, M. J. (2016). Video capsule endoscopy: a tool for the assessment of small bowel transit time.~\textit{Frontiers in medicine},~\textit{3}, 6.

\bibitem{b20} M. Turan, Y. Almalioglu, E. Konukoglu, and M. Sitti, “A deep learning based 6 degree-of-freedom localization method for endoscopic capsule robots,” CoRR, vol. abs/1705.05435, 2017. [Online]. Available: http://arxiv.org/abs/1705.05435

\bibitem{b21} M. Turan, Y. Almalioglu, H. Gilbert, A. E. Sari, U. Soylu, and M. Sitti, “Endo-vmfusenet: Deep visual-magnetic sensor fusion approach for uncalibrated, unsynchronized and asymmetric endoscopic capsule robot localization data,” CoRR, vol. abs/1709.06041, 2017. [Online]. Available: http://arxiv.org/abs/1709.06041

\bibitem{b22} M. Turan, Y. Almalioglu, H. Gilbert, H. Ara´ujo, T. Cemgil, and M. Sitti, “Endosensorfusion: Particle filtering-based multi-sensory data fusion with switching state-space model for endoscopic capsule robots,” CoRR, vol. abs/1709.03401, 2017. [Online]. Available: http://arxiv.org/abs/1709.03401

\bibitem{b23} M. Turan, E. P. Ornek, N. Ibrahimli, C. Giracoglu, Y. Almalioglu, M. F. Yanik, and M. Sitti, “Unsupervised odometry and depth learning for endoscopic capsule robots,” CoRR, vol. abs/1803.01047, 2018. [Online]. Available: http://arxiv.org/abs/1803.01047

\bibitem{b24} M. Turan, Y. Almalioglu, E. P. Ornek, H. Ara´ujo, M. F. Yanik, and M. Sitti, “Magnetic-visual sensor fusion-based dense 3d reconstruction and localization for endoscopic capsule robots,” CoRR, vol. abs/1803.01048, 2018. [Online]. Available: http://arxiv.org/abs/1803.01048

\end{thebibliography}
\end{document}